\documentclass{article}

    \usepackage[preprint]{neurips_2019}

\usepackage[utf8]{inputenc} 
\usepackage[T1]{fontenc}    
\usepackage{hyperref}       
\usepackage{url}            
\usepackage{booktabs}       
\usepackage{amsfonts}       
\usepackage{nicefrac}       
\usepackage{microtype}      
\usepackage{graphicx}
\usepackage{natbib}
\setcitestyle{numbers}

\title{Embedded Encoder-Decoder in Convolutional Networks Towards Explainable AI}

\author{%
 Amirhossein Tavanaei\\
  Department of Data Science and AI\\
  The Procter and Gamble Co.\\
  Cincinnati, Ohio, USA \\
 Email: \texttt{tavanaei.a@pg.com} \\
}

\begin{document}

\maketitle

\begin{abstract}
Understanding intermediate layers of a deep learning model and discovering the driving features of stimuli have attracted much interest, recently. Explainable artificial intelligence (XAI) provides a new way to open an AI black box and makes a transparent and interpretable decision. This paper proposes a new explainable convolutional neural network (XCNN) which represents  important and driving visual features of stimuli in an end-to-end model architecture. This network employs encoder-decoder neural networks in a CNN architecture to represent regions of interest in an image based on its category. The proposed model is trained without localization labels and generates a heat-map as part of the network architecture without extra post-processing steps. The experimental results on the CIFAR-10, Tiny ImageNet, and MNIST datasets showed the success of our algorithm (XCNN) to make CNNs explainable. Based on visual assessment, the proposed model outperforms the current algorithms in class-specific feature representation and interpretable heatmap generation while providing a simple and flexible network architecture. The initial success of this approach warrants further study to enhance weakly supervised localization and semantic segmentation in explainable frameworks. 
\end{abstract}

\section{Introduction}
Convolutional neural networks (CNNs) have shown remarkable performance in differrent areas of pattern recognition, especially computer vision~\cite{lecun2015deep,krizhevsky2012imagenet,rawat2017deep}. To understand how CNNs extract discriminative features from unstructured data, recent studies have visualized the receptive fields (convolution kernels) and feature maps of neural layers to better represent the information flow in a hierarchy of convolutional layers~\cite{qin2018convolutional,zeiler2014visualizing}. However, understanding the intermediate layers of a deep learning model and detecting the driving features of stimuli are challenging and demand new problem statements~\cite{arrieta2020explainable,montavon2018methods}. Explainable artificial intelligence (XAI) helps open the deep learning black box and describe more details about the features extracted in intermediate layers~\cite{gunning2017explainable,adadi2018peeking}. XAI explains the reason behind the prediction by detecting driving features that positively and negatively impact the final neural layer's activation.

More transparent machine learning models such as decision trees are more explainable than high performance, complex deep learning models. Furthermore, reducing the complexity and improving the interpretability of a model may cause significant accuracy drop. The goal is to make the high performance deep learning models more explainable with minimum performance loss and computation cost. A straightforward approach is to improve the deep learning's explainability by combining the feature extractor components of deep learning with transparent models such as rule based approaches~\cite{zilke2016deepred}, boosting trees~\cite{che2016interpretable}, deep K-nearest neighbor (KNN)~\cite{papernot2018deep}, or partitioning the feature space in a treeview model architecture~\cite{thiagarajan2016treeview}. Although, hybrid strategies provide more transparent architectures, they do not offer an end-to-end model architecture with consistent components. 

Decomposing a classifier/predictor function to relevance scores of input dimentions has been studied in~\cite{bach2015pixel} to develop a pixel-wise, layer-wise relevance propagation model. Relevance propagation can also be performed by an approximation method like the Taylor decomposition~\cite{bazen2013taylor,bach2015pixel}. The Taylor decomposition describes the decision made by decomposing the system, $f(x)$, as the sum of relevance scores, $R$~\cite{bazen2013taylor,montavon2018methods}. Thus, the Taylor decomposition can form an XAI formulation by finding the impact of input elements in a deep learning prediction/classification. Based on the divide-and-conquer concept,~\cite{montavon2017explaining} proposed the deep Taylor decomposition method in which the deep learning function can be divided into simpler sub-functions representing relevance scores in consecutive neural layers. They used a first-order Taylor expression around the root, $x_0$, where $w^Tx_0=0$. PatternAttribute proposed by~\cite{kindermans2017learning} extended this model to learn $x_0$ from training data and showed significant improvement in driving features visualization. 

Another common approach for representing deep neural network layers and explaining the model functionality is based on the deconvolutional operations to construct hierarchical image representations~\cite{zeiler2010deconvolutional,zeiler2011adaptive}. This approach has been employed to understand and visualize image representations in neural network layers and to generate class-specific latency maps~\cite{zeiler2014visualizing,mahendran2015understanding}. A convolution-deconvolution network architecture was proposed by~\cite{nguyen2016synthesizing} to generate new images depicting learned features using activation maximization (AM) which identifies the stimulus that maximizes the output neuron's response~\cite{erhan2009visualizing}. In another vein of research, Simonyan et. al.~\cite{simonyan2013deep} showed that the gradient of a neuron's activity with respect to the input is similar to the reconstruction of the corresponding layer. Detecting and representing driving features of stimuli in an XAI framework can be used to develop weakly supervised localization and segmentation~\cite{bazzani2016self,oquab2015object} where the object masks or bounding boxes are not provided. Using the gradient back-propagation through a convolutional network,~\cite{simonyan2013deep} generated saliency maps corresponding to the object's features in an image. The saliency maps in~\cite{simonyan2013deep} were processed to fulfill weakly supervised localization and segmentation tasks. Guided Backprop proposed by~\cite{springenberg2014striving} is similar to the gradient-based saliancy map generators with modifications in using the rectified linear activation functions in backward pass. Regarding the role of explainable convolutional networks in object localization, Zhou et. al.~\cite{zhou2016learning} generated class activity maps (CAM) to indicate the regions of interest using the global average pooling operation over feature maps in the last convolutional layer. Later,~\cite{selvaraju2017grad} proposed the gradient-weighted CAM to use the gradient of target class in the final convolutional layer with a generalized network architecture. The gradient-based approaches mentioned above exhibit the driving features by representing the weight matrices after training. Kindermans et. al.~\cite{kindermans2017learning} introduced a layer-wise back-propagation, named PatternNet, where the information directions are used in the backward pass instead of the weights. 

In this paper, we propose a novel end-to-end explainable CNN, named XCNN, trained in a similar way to the conventional CNNs on classification image datasets without bounding box/segmentation information while no post-processing is required after training. Based on the same criteria,~\cite{zhang2018interpretable} proposed an end-to-end explainable CNN which modifies the traditional CNN architecture and the loss functions to minimize the inter-category entropy and the entropy of the spatial distributions of neural responses. However, our explainable CNN uses the traditional CNN architecture and loss function with more convolutional-deconvolutional layers to develop an encoder-decoder component in the model. Hence, the interpretable component is part of the CNN architecture and generates feature heatmaps based on the images' categories. This encoder-decoder component can be attached to any CNN architecture to enrich its explainability.  

\section{Method}
\begin{figure}
\centering
\includegraphics[scale=.7]{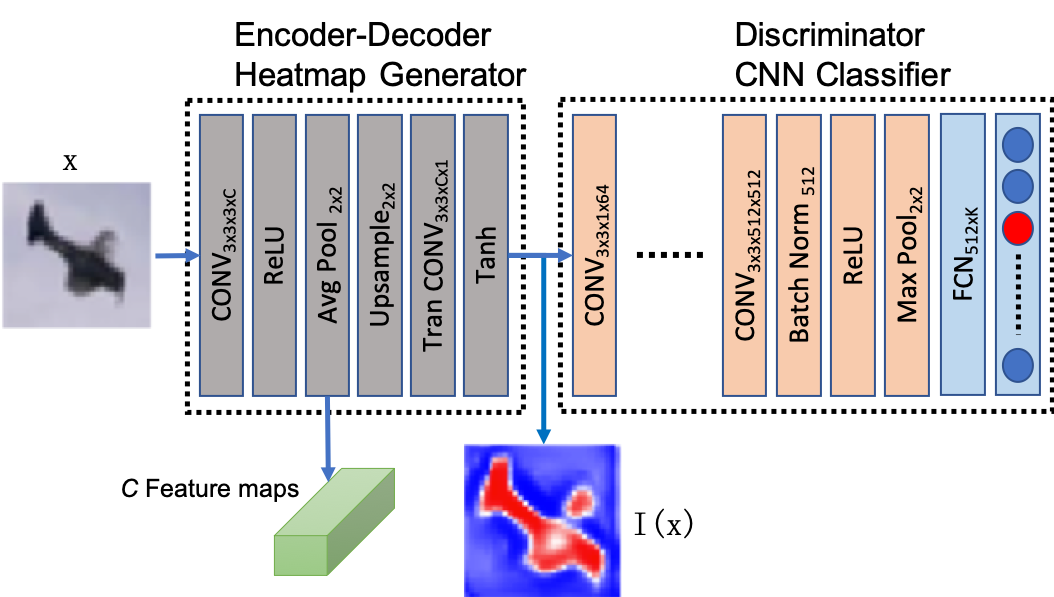}
\caption{The proposed explainable CNN composed of generator and discriminator components. The generator component generates interpretable heatmaps, $I$, in an encoder-decoder neural architecture. The classifier in this figure is VGG16. }
\label{fig:net1}
\end{figure}

The goal of this study is to develop an explainable CNN architecture to be able to extract and depict the driving spatial features of an image, while classifying or predicting the image, in an end-to-end model. This architecture consists of an encoder-decoder component attached to the beginning of a CNN (discriminator) where the encoder-decoder's output is the CNN's input. Figure~\ref{fig:net1} shows the network architecture equipped with the VGG-16 discriminator. 

\subsection{Network Architecture}
This architecture is partially inspired from the conditional and cycle-constraint generative adversarial networks (GANs)~\cite{goodfellow2016nips,isola2017image,zhu2017unpaired} where the generator component generates new images based on the training dataset of real images and also the features extracted from the images of another domain (let's say input images). The input image set specifies the features that are transferred to the training dataset's domain. The encoder-decoder component (generator) of the XCNN generates a single-channel image (heatmap) that is fed into a discriminator/classifier, $D$. Eq.~\ref{eq1} formulates the heatmap ($I(x)$).
\begin{equation}
I(x) = \mathrm{Tanh} \big (\mathrm{Decode} (\mathrm{Encode}(x) \big )
\label{eq1}
\end{equation}
The downsampling operation in the generator part is implemented by the average pooling to smoothly extract features and to preserve localization information. The `Tanh(.)' activation function in the last layer of the generator normalizes the heatmap in the range [-1,1] and submits positive and negative values to the discriminator component. 

\subsection{Information Flow}
The generator loss in a GAN is defined by:
\begin{equation}
G_{loss} = \log \big (1- D\big ( G(z)\big ) \big )
\label{eq2}
\end{equation}
Which shows how the discriminator ($D$) is fooled by the fake images ($G(z)$) generated by $G$ based on the input $z$. The probability distribution of the generated images gets closer to the probability distribution of training dataset during the learning process. However, in our explainable CNN, the goal is to extract discriminative features that distinguish different image categories. Hence, a classification loss function is used in this algorithm as follows:
\begin{equation}
loss = \log \big ( D \big ( I(x) \big ) \big )
\label{eq3}
\end{equation}
Where, $I(x)$ identifies the heatmap generated by the encoder-decoder (generator) component (Eq.~\ref{eq1}). In the other words, the encoder part extracts important features and the decoder part shows the driving features in a new format preserving the spatial information of the driving features in a classification task. According to the information theory in neural networks~\cite{tishby2015deep}, in an intermediate layer, more uncertainty is removed from the input, $x$, if the output, $y$, is known and the mutual information between an intermediate layer (e.g. the heatmap layer) and the output ($y$) is larger than the mutual information between the input ($x$) and the output ($y$).
\begin{equation}
\mathrm{Inf}(I(x),y)>\mathrm{Inf}(x,y)
\label{eq:4}
\end{equation}
\begin{equation}
\mathrm{Inf}(x,y) = H(x) - H(x|y), \ \ H: Entropy
\end{equation}
Therefore, we obtain more mutual information between $I(x)$ and $y$ while describing the classifier's ($D$) decision in a heatmap by localizing the important pixels of the image $x$.

\section{Experiments and Results}
To evaluate the proposed algorithm, the MNIST~\cite{lecun1998mnist}, CIFAR-10~\cite{krizhevsky2009learning}, and Tiny ImageNet~\cite{Le2015TinyIV} datasets were used. The experiments on the MNIST dataset demonstrate the primary success of the XCNN in extracting interpretable heatmaps and warrants further experiments on larger datasets. The results of the model on the two other datasets are compared with the state of the art interpretable CNNs and saliency map generators. The source codes are available at \url{https://github.com/tavanaei/ExplainableCNN}.

\subsection{MNIST Dataset}
\begin{figure}
\centering
\includegraphics[scale=.45]{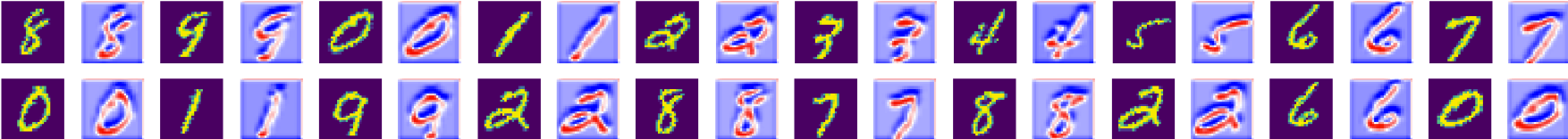}
\caption{Randomly selected MNIST digits classified by the proposed explainable CNN. The blue-red images demonstrate the heatmaps generated in the network.}
\label{fig:mnist}
\end{figure}

\begin{figure}
\centering
\includegraphics[scale=.474]{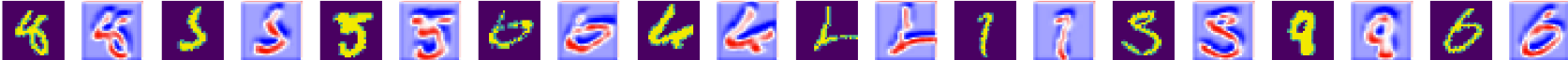}
\caption{Incorrectly classified MNIST digits. The actual labels are `8 5 5 6 6 2 1 3 8 6' and the predicted labels are `4 3 3 0 4 6 7 5 9 0', from left to right.}
\label{fig:mnist2}
\end{figure}

The MNIST dataset includes $70,000$ ($60$k training samples, $10$k testing samples) $28\times28$ gray-scale images of handwritten digits. The XCNN developed for this experiment consists of a generator component including 32 feature maps in its encoder\footnote{The generator's details: the convolutional blocks explained in Fig.~\ref{fig:net1} with 3 input channels, 32 features maps (kernels) for the next block, and 1 feature map for the last block.}; and 1-16-32-64 convolutional kernels (with $3\times 3$ kernel size) followed by a 576-10 fully connected layer\footnote{A fully connected layer with 576 input neurons and 10 output neurons.} for the discriminator component. The accuracy rate of the XCNN was almost the same as the accuracy rate of the conventional CNN used in the discriminator component. We expect larger accuracy drops for more complex datasets as a small information loss is experienced by reducing the input channels to one in the heatmap. Figure~\ref{fig:mnist} demonstrates randomly selected MNIST digits from the test set and their corresponding heatmaps ($I(x)$) generated during the classification task. Figure~\ref{fig:mnist2} shows ten incorrectly classified images and explains how driving features pick the predicted class. 

\subsection{CIFAR-10 Dataset}
\begin{figure}
\centering
\includegraphics[scale=.9]{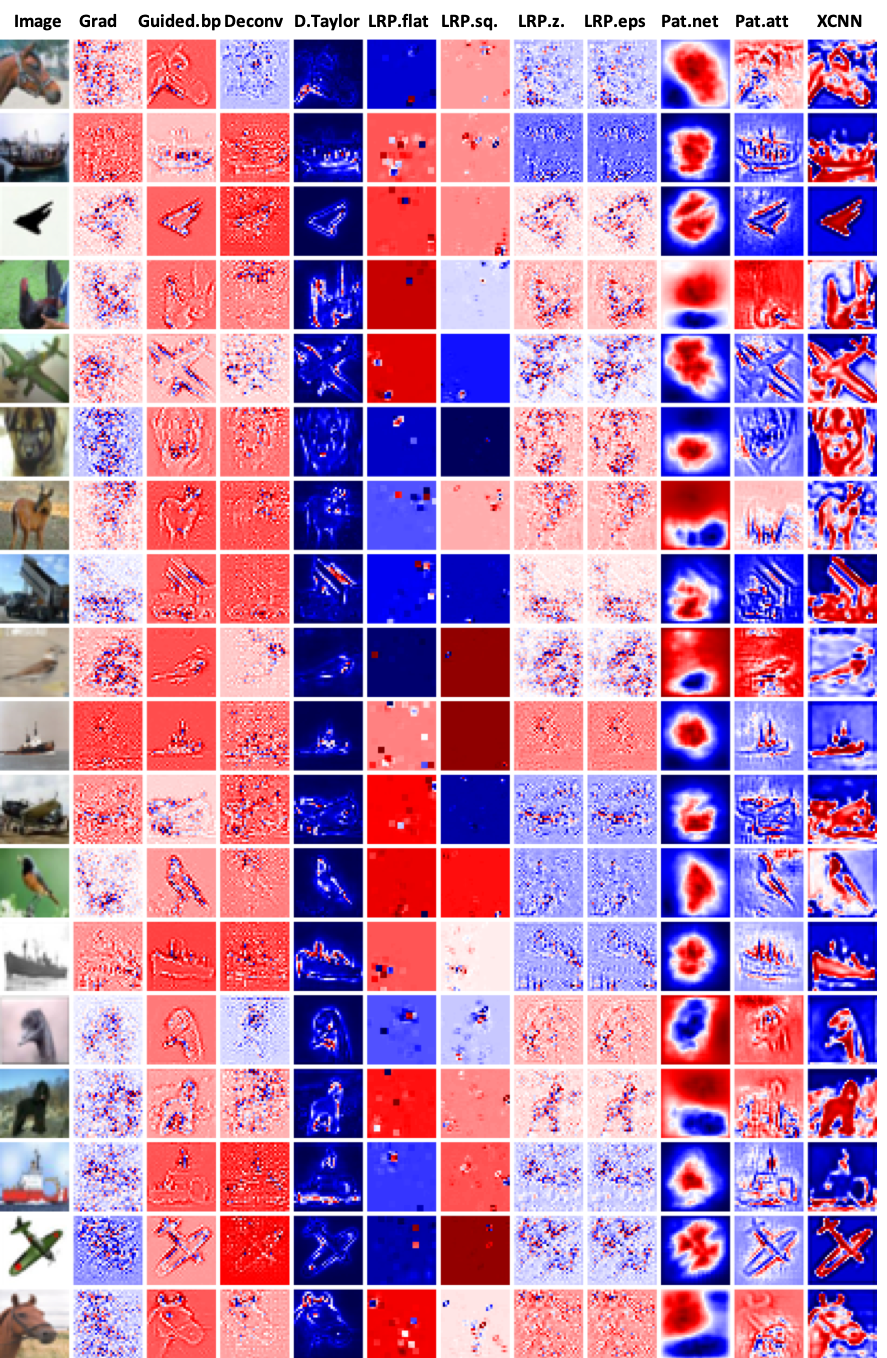}
\caption{Test samples from the CIFAR-10 dataset. The heatmaps are generated based on the VGG-16 model trained on the CIFAR-10 dataset by the `Innvestigate' tool~\cite{alber2019innvestigate}. The algorithms from left to right are Gradient~\cite{simonyan2013deep}, Guided Backprop~\cite{springenberg2014striving}, Deconvolution~\cite{zeiler2014visualizing}, Deep Taylor Decomposition~\cite{montavon2017explaining}, Layer-wise Relevance Propagation (LRP)~\cite{bach2015pixel}, PatternNet and PatternAttribute~\cite{kindermans2017learning}, and our algorithm (XCNN). }
\label{fig:cifar}
\end{figure}

The second experiment involves the CIFAR-10 dataset including $60,000$ images of size $32\times 32$ labeled with ten object categories. The XCNN consists of a generator with 3-128-1 convolutional kernels (feature maps), and the VGG-16 architecture with the input channel of one as discriminator. Figure~\ref{fig:cifar} compares our algorithm with the baselines and recent explainable models and saliency map generators. The XCNN, deep Taylor decomposition, and PatternAttribute methods depict the best object localization and driving feature representation among other models. 

\subsection{Tiny ImageNet Dataset}
\begin{figure}
\centering
\includegraphics[scale=.9]{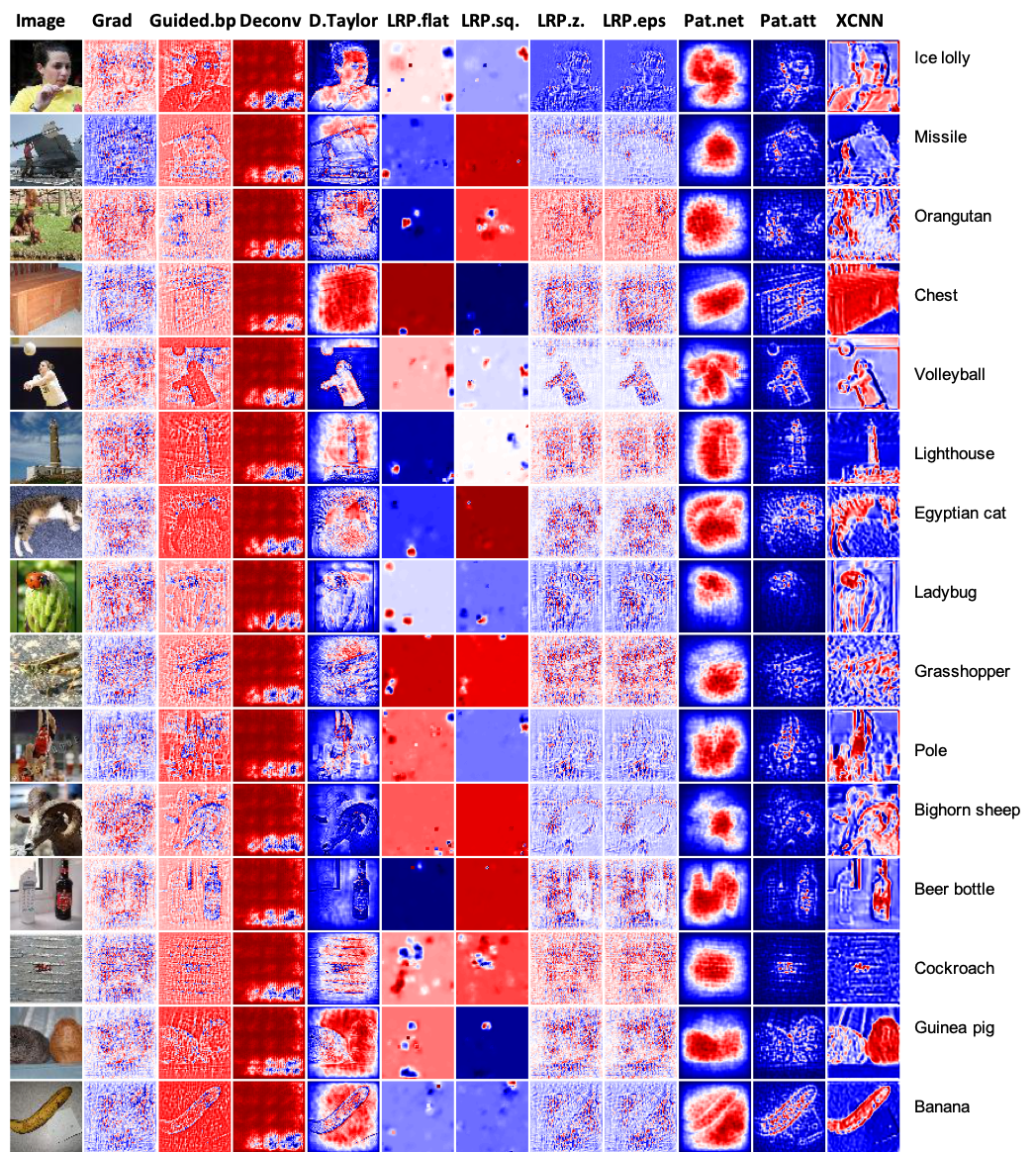}
\caption{Test samples from the Tiny ImageNet dataset. The heatmaps are generated based on the VGG-16 model trained on the Tiny ImageNet dataset by the `Innvestigate' tool~\cite{alber2019innvestigate}. The algorithms from left to right are Gradient~\cite{simonyan2013deep}, Guided Backprop~\cite{springenberg2014striving}, Deconvolution~\cite{zeiler2014visualizing}, Deep Taylor Decomposition~\cite{montavon2017explaining}, layer-wise relevance propagation (LRP)~\cite{bach2015pixel}, PatternNet and PatternAttribute~\cite{kindermans2017learning}, and our algorithm (XCNN). Class labels are shown in the last column.}
\label{fig:imagenet}
\end{figure}
In this experiment, the MicroImageNet challenge's (Tiny ImageNet) dataset is used. This dataset contains $100000$ images of 200 classes (500 for each class) downsized to $64\times 64$ colored images.  Similar to the previous experiment, the  XCNN consists of a generator with 3-128-1 convolutional kernels, and the VGG-16 architecture with the input channel of one as discriminator. Figure~\ref{fig:imagenet} shows that the XCNN, deep Taylor decomposition, and PatternAttribute methods depict the best object localization and class-specific saliency maps. In the most cases, XCNN's heatmaps represent more details and better indicate the target object's pixels than the other models. A number of samples in Figure~\ref{fig:imagenet} are complex and include non-target (but similar to the target) objects in the image as well (e.g. a phone next to the beer bottle). The class labels are shown in the rightmost column to see how different models are depicting the target object's features. The XCNN and PatternAttribute algorithms perform better than the other approaches in picking the target class.

\subsection{Discussions}
The generator component of the XCNN architecture can be utilized in any CNN architecture (discriminator) as it does not change or use the discriminator. The extra layers created in this architecture slightly increase the time and space complexity of the classifier while generating heatmaps in addition to the class prediction in an end-to-end structure. However, the end-to-end architecture of the XCNN does not need extra memory for keeping the gradient of signal in the test/prediction phase. Additionally, generating the heatmaps is part of the classification task and it does not need further information processing.

\begin{figure}
\centering
\includegraphics[scale=.95]{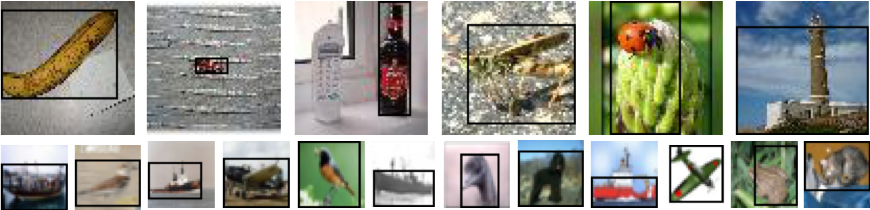}
\caption{Object localization test examples. The heatmaps generated by the XCNN are thresholded (threshold=0.5) and the largest contours indicate the bounding boxes. \textit{This figure only shows an example of possible future use cases of the proposed explainable architecture.}}
\label{fig:segment}
\end{figure}

The contrast shown in the XCNN heatmaps supports weakly supervised pixel-wise segmentation and localization after processing the generated heatmaps. Figure~\ref{fig:segment} shows a number of object localization examples that are obtained simply by thresholding the generated heatmaps. Only thresholding the heatmap is not performing well for general cases. Hence, to improve the localization and segmentation results, a better strategy such as using the GraphCut~\cite{boykov2001interactive} as explained in~\cite{simonyan2013deep} is required\footnote{Will be addressed in our future study}. 

\begin{figure}
\includegraphics[scale=.65]{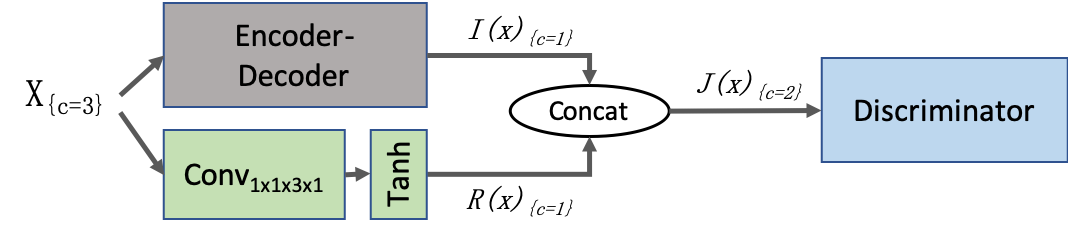}
\centering
\caption{The modified XCNN using $1\times 1$ convolution next to the generator component for improving the accuracy rate. $C$ indicates the number of features maps (channels).}
\label{fig:res}
\end{figure}

\begin{figure}
\includegraphics[scale=.77]{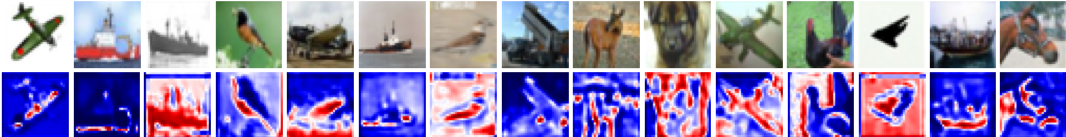}
\centering
\caption{Examples of interpretable maps ($I(x)$) generated by the modified XCNN architecture on the CIFAR-10 to improve the classification accuracy.}
\label{fig:res_map}
\end{figure}

Regarding the classification accuracy, the classifier's input is a heatmap with one channel summarizing the visual features of an input image so that this summarization may cause accuracy reduction in classifying complex image datasets. 
The training and validation accuracy rates of the XCNN on the CIFAR-10 and the Tiny ImageNet were [0.71\% \& 2.80\%] and [0.45\% \& 2.77\%] less than the original VGG-16, respectively. Although this range of accuracy drop is expected, the accuracy can be improved by adding a $Conv_{1\times 1}$ next to the generator component as shown in Figure~\ref{fig:res}. The modified architecture improved the accuracy rate (about 2\% validation accuracy improvement on the CIFAR-10); however, the quality of the interpretable heatmaps ($I(x)$) were tainted as shown in Figure~\ref{fig:res_map}. As our focus in this study is to improve the explainability of convolutional networks, the accuracy loss in the original XCNN is acceptable. Improving the accuracy and using other discriminator architectures are our future work plan.

\section{Conclusion}
Explainability of high performance deep learning models is an important and challenging problem in different research areas, especially computer vision. This paper proposed a new explainable convolutional neural network (XCNN) to represent the driving visual features of stimuli in an end-to-end network architecture. The proposed model consists of two consecutive components: 1) a heatmap generator built by encoder-decoder neural layers and 2) a CNN classifier. Experimental results on the MNIST, CIFAR-10, and Tiny ImageNet datasets showed interpretable heatmaps that visually outperformed the state of the art explainable networks and saliency map generators while offering a simple architecture that can be reapplied to any CNN classifier. 

The success of the XCNN in discovering main visual features and generating heatmaps warrants further study to utilize this network architecture and generated heatmaps in other computer vision research categories. Our future work seeks to employ the XCNN to develop new weakly supervised localization, semantic segmentation, and object tracking in video streams.

\subsubsection*{Acknowledgments}
Special thanks to Dr. Venugopal Vasudevan and Dr. Kelly Anderson at P\&G for their constructive comments and support.


\bibliographystyle{unsrtnat} 
\small
\bibliography{amir}

\end{document}